\documentclass[a4paper, 10pt, conference]{ieeeconf}

\newif\ifoagmfinalcopy

\oagmfinalcopytrue  

\IEEEoverridecommandlockouts                       

\usepackage{graphics} 
\usepackage{epsfig} 
\usepackage{mathptmx} 
\usepackage{times} 
\usepackage{amsmath} 
\usepackage{amssymb}  

\usepackage{lineno}
\usepackage{tikzpagenodes}
\usepackage{background}

\usepackage{subcaption}
\usepackage{booktabs}
\usepackage{siunitx}
\usepackage{url}

\usepackage{fancyhdr}

\ifoagmfinalcopy
\backgroundsetup{color=white}
\else

\linenumbers

\newcommand{\MyOAGMConfidentialLogo}{
\begin{tikzpicture}[remember picture,overlay]
\node[align=center,text=blue] at ([yshift=1em]current page text area.north) {\Large \#\#\# ARW/OAGM 2021 SUBMISSION: CONFIDENTIAL REVIEW COPY \#\#\#};
\end{tikzpicture}%
}

\SetBgContents{\MyOAGMConfidentialLogo}
\SetBgPosition{current page.north west}
\SetBgOpacity{0.5}
\SetBgAngle{0.0}
\SetBgScale{1.0}

\fi

\title{\LARGE \bf
SliTraNet: Automatic Detection of Slide Transitions in Lecture Videos using Convolutional Neural Networks}

\ifoagmfinalcopy
\author{Aline Sindel$^{1}$, Abner Hernandez$^{1}$, Seung Hee Yang$^{2}$, Vincent Christlein$^{1}$ and Andreas Maier$^{1}$
\thanks{This work was supported by the project ``MEOW'' funded by ``DAAD IP Digital''}
\thanks{$^{1}$Pattern Recognition Lab, Friedrich-Alexander-Universität Erlangen-Nürnberg (FAU), 91058 Erlangen, Germany {\tt\small \{aline.sindel, abner.hernandez, vincent.christlein, andreas.maier\}@fau.de}}%
\thanks{$^{2}$Department of Artificial Intelligence in Biomedical Engineering, Friedrich-Alexander-Universität Erlangen-Nürnberg (FAU), 91058 Erlangen, Germany {\tt\small 
seung.hee.yang@fau.de}}%
}
\else
\author{Anon, Ymous}
\fi

\begin{document}

\maketitle
\thispagestyle{fancy}
\fancyhead{}
\lhead{}
\vspace{-0.5pt}
\lfoot{\scriptsize{Accepted to the Austrian Association for Pattern Recognition (OAGM) Workshop 2021 - Computer Vision and Pattern Analysis Across Domains.
}}
\cfoot{}
\rfoot{}

\newcommand{\source}[1]{\vspace{-5pt} \caption*{\scriptsize{{#1}}} }
\newcommand{\sourcesmall}[1]{\vspace{-5pt} \hfill \scriptsize{{#1}} }

\begin{abstract}
With the increasing number of online learning material in the web, search for specific content in lecture videos can be time consuming. Therefore, automatic slide extraction from the lecture videos can be helpful to give a brief overview of the main content and to support the students in their studies.
For this task, we propose a deep learning method to detect slide transitions in lectures videos. We first process each frame of the video by a heuristic-based approach using a 2-D convolutional neural network to predict transition candidates. Then, we increase the complexity by employing two 3-D convolutional neural networks to refine the transition candidates.
Evaluation results demonstrate the effectiveness of our method in finding slide transitions.
\end{abstract}

\section{INTRODUCTION}
Nowadays, there is a huge number of online learning material available to students and researchers. Lecture videos uploaded by the universities to video sharing platforms such as YouTube or to in-build video platforms are accessible from anywhere and at any time. The high amount of video material makes it tedious for the user to search for specific content by browsing through the individual videos. Hence, video summarization can help to quickly grasp the overview of the lecture video. This can be done by the automatic detection of slide transitions to extract the slide and time stamp at each slide change. Automatic detection of slide transitions can also support the lecturer in creating lecture notes. In combination with the audio transcript of the lecture video, the extracted slides can be automatically inserted into the audio text based on their time stamp. 
 For instance, the free video-to-blog post conversion software AutoBlog~\cite{MaierA2020} automatically extracts the transcript of a lecture video to generate a blog post~\cite{HernandezA2021}. So far, the slides are manually inserted into the blog text. However, using our slide transition detection method, the software could be extended.  
 
The variety in the types of lecture videos makes the task challenging. For example, the lecture slides can be full screen with the lecturer screen inserted as a small window on top, or the lecture slides can be depicted next to the view of the lecturer. 
Further, memes (e.g. animations and short videos) to illustrate the lecture content can be inserted into the lecture video. Memes and the actual slides can have very similar frames from the style and color distribution. Thus, lecture videos that not only contain the slides and the speaker's view, but also these meme videos make the task even more difficult.

\begin{figure}[t] 
\centering
\includegraphics[width=8cm]{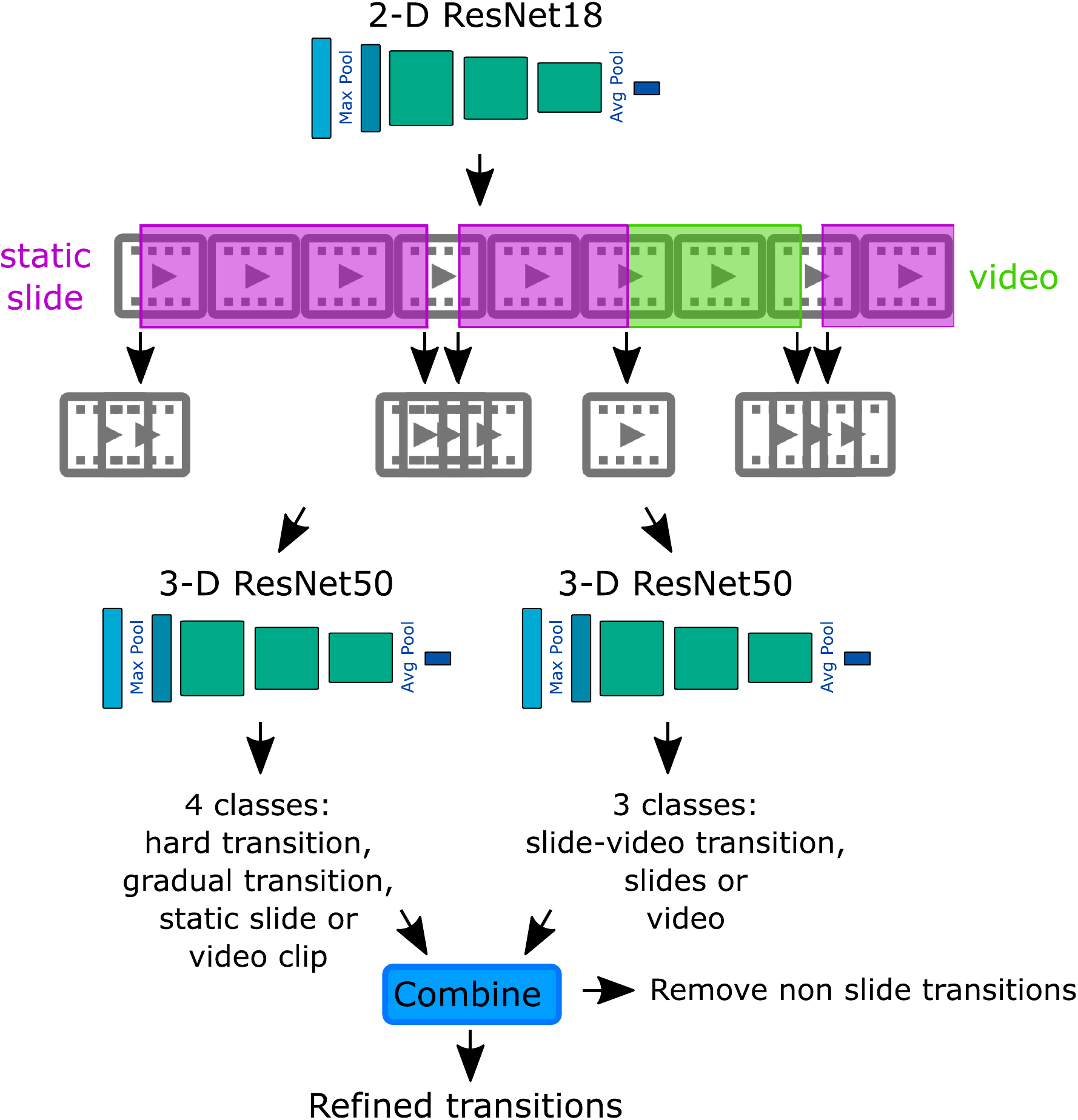}
\caption{Overview of our SliTraNet for slide transition detection: First, we predict initial slide-slide or slide-video transition candidates by comparing each frame (cropped to the slide content) to its respective anchor frame using a \mbox{2-D ResNet}. At the transition candidate positions, we extract overlapping video clips with a length of eight frames from the cropped video and the raw video. Two 3-D ResNets have been trained to extract spatio-temporal features to classify the cropped video clips into hard or gradual transitions, static slides or video sequences and the raw video clips into slide-video transitions, slide sequences or video sequences. Lastly, we combine the class predictions of both 3-D ResNets to exclude transitions mutually classified as video sequence.}
\label{fig:inference}
\end{figure}

In this paper, we propose a deep learning method for the detection of slide transitions in lecture videos, which we train and test on a dataset that contains video sequences of lectures with slides, speaker views, and memes. 
To detect the slide transitions, we present a multi-step approach. First, we predict initial transition candidates by inserting a \mbox{2-D} convolutional neural network (CNN) into a heuristic-based approach. Then, we extract spatio-temporal features at the candidate positions using two 3-D CNNs to exclude transitions that were classified as video sequences.

\section{RELATED WORK}
This section summarizes related works in the field of slide transition detection, scene boundary detection, and video thumbnail selection.
\subsection{Slide Transition Detection}
Traditional approaches to slide detection focus on low-level features to measure the similarity across adjacent frames. For example, the maximum peak of the color histogram and difference in entropy for horizontal lines were used to detect slide changes in~\cite{emeeting}. Often the use of histograms for slide detection is supplemented with other algorithms to detect features such as faces, or text~\cite{ma2012,zhao2018}. Similarly in~\cite{eruvaram2020}, histograms are utilized for shot boundary detection as part of a larger scheme involving shot classification, slide region detection, and slide transition detection.

The variance in image scaling and rotation can be handled by the Scale Invariant feature transform (SIFT) algorithm. This approach detects slide transitions when the SIFT similarity is under a defined threshold. Features extracted using the SIFT algorithm have shown good slide detection accuracy rates in~\cite{jeong2015,Mavlankar2010} and with slide alignment~\cite{wang2009}. SIFT features can also be used with sparse time-varying graphs~\cite{liu2017}, where the graph models slide transitions. The temporal modeling of slide transitions can also be conducted using a Hidden Markov Model (HMM), where the states of the model correspond to an individual slide~\cite{gigonzac2007,schroth2011,fan2007,fan2011}. The likelihood of the states are computed with a correlation measure and the most probable sequence of slides is calculated using the Viterbi algorithm. 

The current study approaches the slide transition detection problem by using 3-D CNNs which can learn spatio-temporal features that are useful for detecting slide transitions. However, the training time and memory consumption can be problematic. Therefore, Residual Networks (ResNet~\cite{HeK2016}) have been suggested by~\cite{liu2019} for this task. They propose a novel residual block that contains an extra 1×1 3-D convolutional layer to the shortcut connection layer. 
They show better results for ResNet compared to the traditional slide transition approaches on their to 6 frames per second temporally downsampled dataset.
In~\cite{guan201}, a Dual Path Network (DPN)~\cite{ChenY2017} that combines both ResNeXt and DenseNet is proposed. Further, they introduce a Convolutional Block Attention module to their network that sequentially infers a 1-D channel attention map, followed by a 2-D spatial attention map, and lastly a \mbox{1-D} time attention map. Further improvements in the $F_1$-score were obtained compared to traditional approaches or with ResNets alone.

\subsection{Scene Boundary Detection}
A related field of work is scene boundary detection or shot boundary detection (SBD)~\cite{jiang1998}. Traditionally, SBD relied on the same low-level features such as histograms. However, the issue of detecting changes is complex and requires attention to the variability of transitions. Detecting gradual transitions is a particularly difficult problem and recent studies on SBD now take into consideration the presence of sharp cut transitions and gradual transitions. For example, a 3-D CNN-based model from~\cite{hassanien2017} was combined with an SVM classifier to label frames as being either normal, a gradual transition, or a sharp transition. In~\cite{liang2017}, both types of transitions are detected by separate 3-D CNNs. A similar approach using deep CNNs was taken by~\cite{wang2021} where SBD was implemented via a three stage process; candidate detection, cut transition detection, and gradual transition detection. TransNet~\cite{souvcek2019} and TransNet2~\cite{souvcek2020} use Dilated DCNNs to detect sharp and gradual transitions. 

\subsection{Video Thumbnail Selection}
Another related area is video thumbnail selection, which summarizes the video content by selecting a representative frame as the thumbnail.
To extract the representative frames, learning-based approaches have been proposed that take the user's perspective selection of representative frames into account~\cite{KangHW2005,LuoJ2009}. Based on visual features, the videos are classified according to image quality, visual details, user attention, and display duration~\cite{KangHW2005}, or different types of camera motion~\cite{LuoJ2009}.
Approaches also combine the visual content with side semantic information such as the title or transcript for query-dependent thumbnail selection~\cite{LiuW2015} or to visually enrich the thumbnail with keywords~\cite{ZhaoB2017}.

\begin{figure*}[t] 
\begin{center}
\begin{subfigure}[b]{.39\linewidth}
\centerline{\includegraphics[width=7cm]{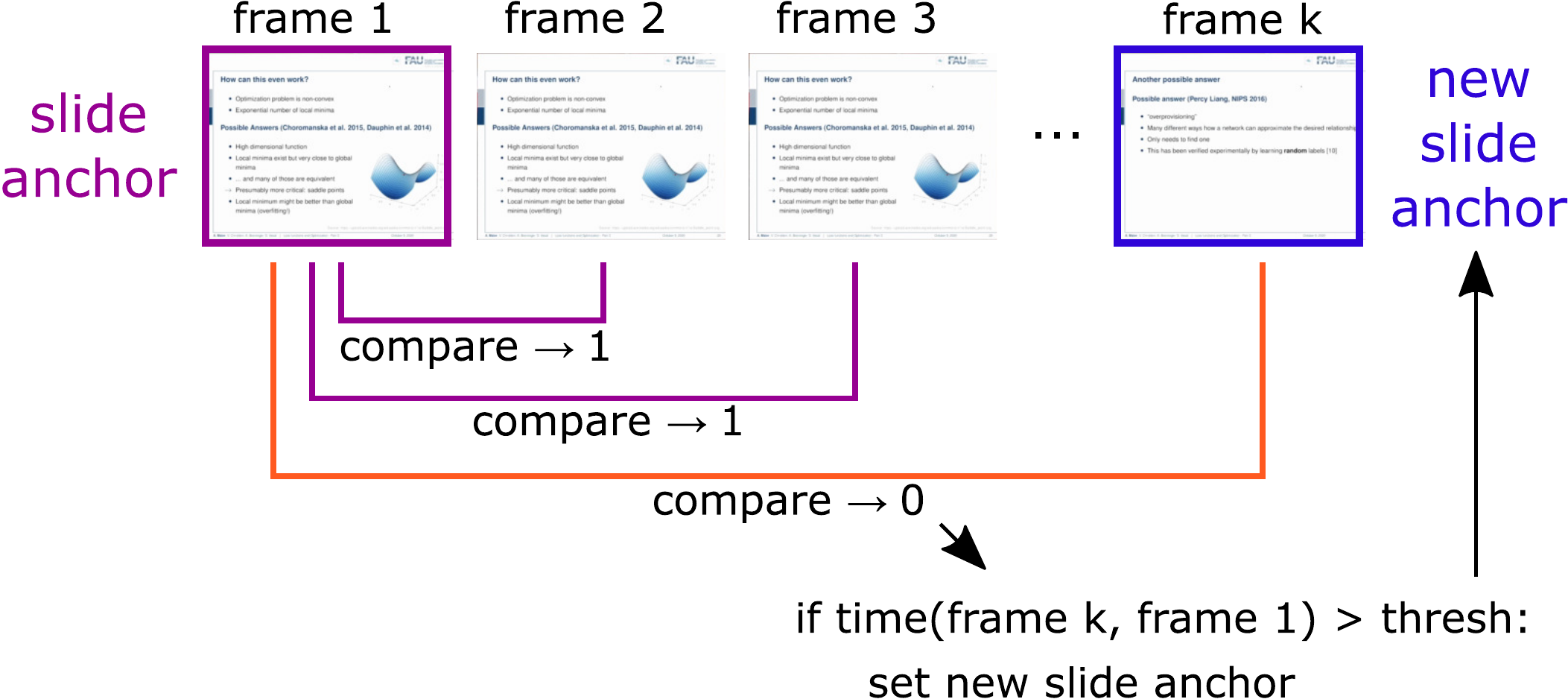}}
\caption{~Detection of a static slide sequence}\label{fig:compare_2d_static_slide}
\end{subfigure}
\hfill
\begin{subfigure}[b]{.59\linewidth}
\centerline{\includegraphics[width=9cm]{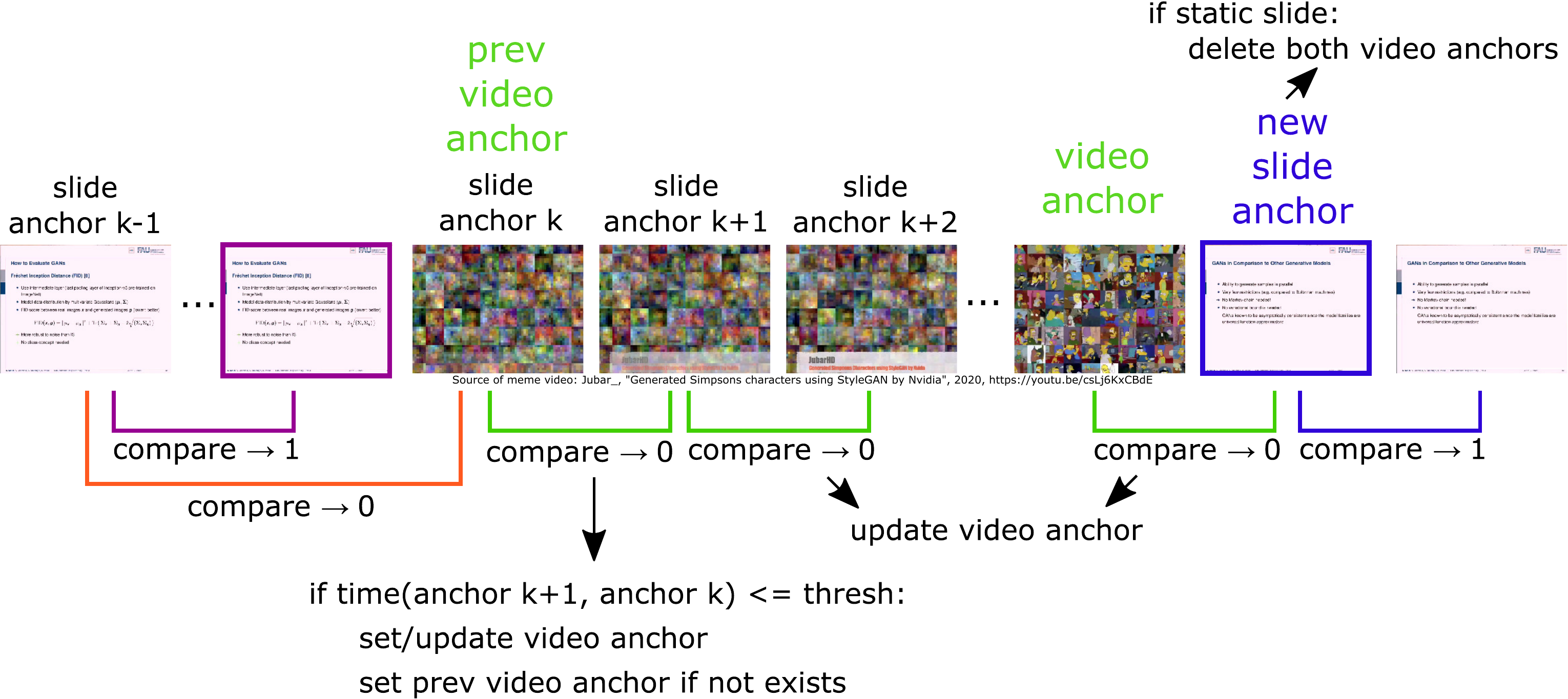}}
\caption{~Detection of a video frame sequence}\label{fig:compare_2d_video}
\end{subfigure}
\end{center}
\caption{Comparison to anchor frame using a neural network to detect static slide sequences and video sequences.} 
\label{fig:compare_2d}
\end{figure*}

\section{METHODOLOGY}
In this section, our method for slide transition detection is presented. We describe the network architectures and introduce training and inference of the different parts of the pipeline.

\subsection{Overview of SliTraNet} 
SliTraNet is composed of three convolutional neural networks, which are all separately trained for the three different tasks and combined for inference, see Fig.~\ref{fig:inference}. 
We process the complete data once by applying a 2-D ResNet18~\cite{HeK2016} to pairs of each frame with its anchor frame resulting in initial slide-slide or slide-video transition candidates. For the refinement step, we increase the complexity of the networks by using two 3-D ResNet50s and apply these to video clips extracted from the transition candidate positions. 
A short video clip of eight frames can contain a sequence with one hard transition, a gradual transition, a static sequence of the same slide or a sequence of video frames, such as a short animation, a speaker view, or a meme.
We train one \mbox{3-D ResNet} for these four classes and another 3-D ResNet to distinguish slide-video transitions, slide sequences, and video sequences. Based on the class predictions, we exclude transition candidates that were classified as video sequence by both networks.

\subsection{Initial Transition Candidates Estimation}
We train a 2-D ResNet18 for the discrimination task whether two images are from the same slide (class 1) or not (class 0). For this task, we concatenate both images along the color channel dimension to obtain 6 channels for RGB input or 2 channels for grayscale input and modify the input channels of the ResNet18~\cite{HeK2016} architecture accordingly. For training, we generate the same number of positive and negative pairs. For the negative pairs, we first select frames from the neighboring slides for each slide and then fill the rest with randomly chosen frames that have a different slide id. For the optimization, we employ the binary cross-entropy loss.

To predict the transition candidates, we plug the neural network into a heuristic-based approach, as illustrated in Fig.~\ref{fig:compare_2d_static_slide} and~\ref{fig:compare_2d_video}. We compare each frame to an anchor frame by the neural network to search for static slides (Fig.~\ref{fig:compare_2d_static_slide}). As long as both frames are classified as the same, we keep the anchor and as soon as the two frames are classified as different, we set the anchor to the current frame. A static slide is detected if the time, measured in number of frames, is higher than a threshold. This general idea is borrowed from Perelman~\cite{PerelmanD2020}, which uses the absolute difference of the blurred grayscale versions of the frames. Since the lecture videos also contain video sequences without slides, we extended the approach further by adding two video anchors, see Fig.~\ref{fig:compare_2d_video}. If a frame difference is detected by the neural network and the time from the current frame $k$ to the anchor $k-1$ is smaller or equal to the threshold, the video anchor and previous video anchor are set to the current frame $k$. As long as the frames are not classified as the same, the video anchor is updated. After the next static slide sequence is detected, a video sequence is recorded from the previous video anchor to the video anchor and both anchors are deleted. The slide-slide and slide-video transition candidates are determined from the detected static slide and video sequences.

\begin{figure*}[t] 
\centering
\includegraphics[width=14cm]{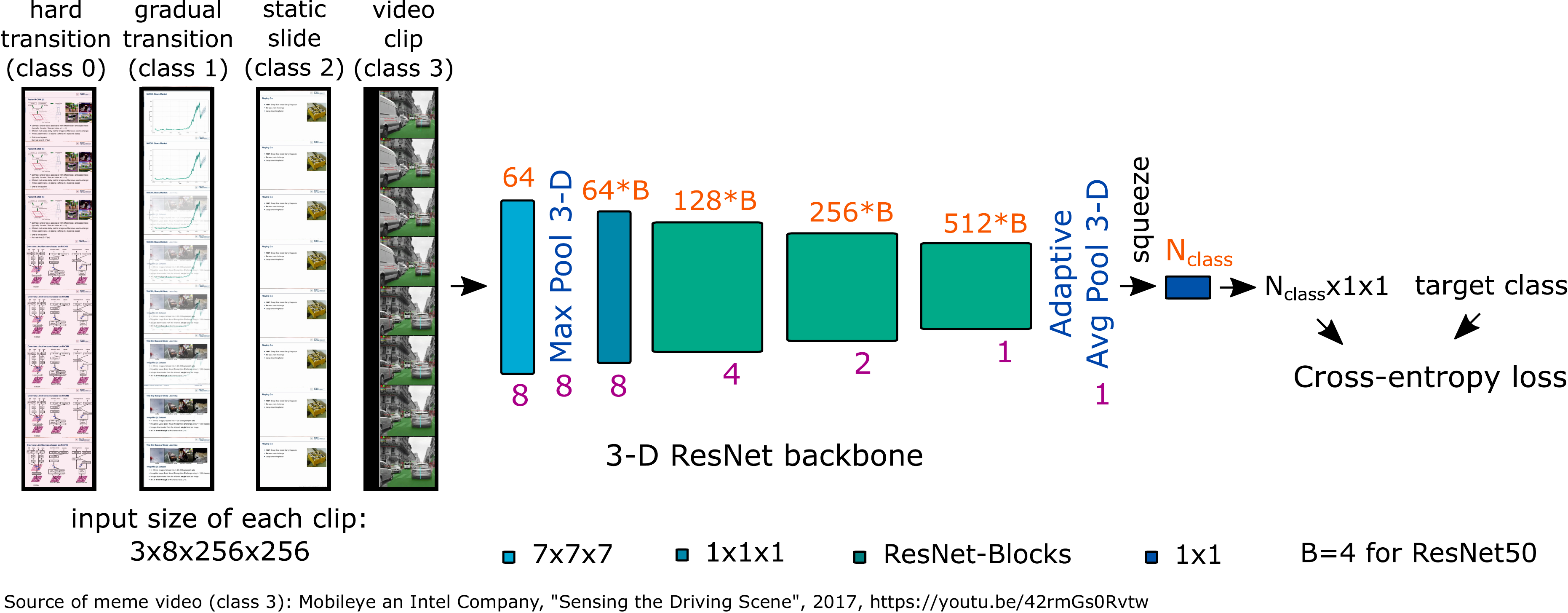}
\caption{Training of 3-D ResNet50 for the multi-class classification task: detection of hard transitions, gradual transitions, static slides, and videos. For each input clip one class is predicted. The numbers in orange indicate the number of output feature maps of each convolutional layer or block and the numbers in purple denote the output dimension of the temporal dimension.} %
\label{fig:ResNet3d}

\end{figure*}

\subsection{Transition Candidates Refinement}
Since the video sequences can also contain static frames that might be classified as static slides using the deep-heuristic-based approach, a refinement step is necessary to reduce the number of false positives. 
To better exploit the spatio-temporal character of the video, we train a \mbox{3-D ResNet50} using cross-entropy loss for the multi-task classification problem that assigns a short video clip of eight frames to one of the classes: hard transition, gradual transition, static slide and video. The network architecture is depicted in Fig.~\ref{fig:ResNet3d}, which is slightly adapted from the 3-D ResNet backbone in \cite{KopuOklu2019}. For the initial layers and 3-D max pooling, we modified the strides of the temporal dimension to be $1$ to only reduce the spatial dimensions. 

\begin{figure}[t] 
\centering
\includegraphics[width=7.5cm]{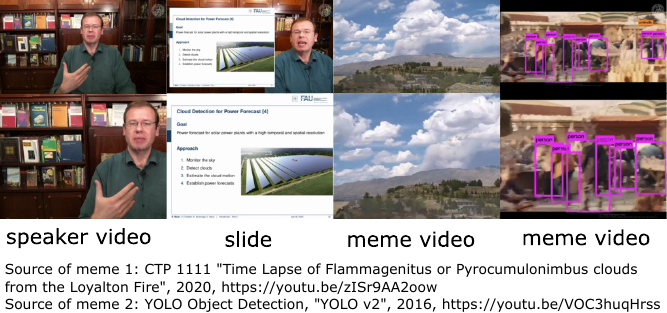}
\caption{Frames of the lecture video dataset. Top row: raw video frames, bottom row: cropped frames.}
\label{fig:dataset}
\end{figure}

The slides of lecture videos are not necessarily filling the full screen, but can be placed on top of some background. In our particular lecture video dataset, the memes, animations and speaker video sequences are full screen in contrast to the slides, see Fig.~\ref{fig:dataset}. Using this knowledge, we use the raw video input to train our second 3-D ResNet50 to classify the short clip into slide-video transitions, slide sequences or video sequences.

For training both networks, we extract video clips at striking positions such as placing the middle of the clip (plus minus one frame) at the position of the hard transition, the begin, middle and end of the gradual transition and in the middle of a static slide sequence or at some equally spaced positions within the video sequence. For the second task, the slide-video transitions occur only rarely in the dataset in comparison to slide sequences or video sequences. Hence, we use the weighted cross-entropy loss to account for the frequency of the classes.

During inference, we use the predictions of the deep-heuristic-based approach to extract the video clips to feed them to the 3-D ResNets and based on the output of both networks, we filter out slide transition candidates that were classified to be video sequences by both networks.

\section{EXPERIMENTS AND RESULTS}
In this section, we describe our dataset and measure the performance of our method.

\subsection{Lecture Video Dataset}
The dataset comprises a subset of lecture videos from two courses in the field of deep learning and medical image processing of the Pattern Recognition Lab, FAU Erlangen-Nuremberg. The videos are recorded in Full HD with $25$ frames per second and range between a duration of $6$ to $33$ min. The slides of one course are in the format $4:3$ and of the other in $16:9$. The dataset is split into $12$ videos for training, $4$ for validation and $14$ for testing. To feed the data to the network, the frames are cropped to the content of the slides except for the video-slide differentiation task (see Fig.~\ref{fig:dataset}) and for all tasks are scaled to a maximum length of $256$ and filled up with zero padding to a patch size of $256 \times 256$.
The ground truth slide transitions were obtained semi-automatically. Based on the difference of the frames, static slides were roughly detected and were manually corrected at frame level and split into hard and gradual transitions.

\subsection{Implementation Details}
We trained all networks from scratch for $100$ epochs with early stopping using the following training parameters: learning rate $\eta=2\cdot10^{-4}$, linear decay to $0$ starting at epoch $50$ for 2-D ResNet18 and $60$ for 3-D ResNet50, Adam solver, momentum $(0.9, 0.999)$, batch size of $64$ for 2-D ResNet18 and of $32$ for 3-D ResNet50 for training and validation and online data augmentation for the training data split (color jittering, horizontal flipping, color inversion, Gaussian blurring with kernel size in range 1 to 21, reversed ordering of the clips and one frame offsets at clip extraction). 
For inference, the threshold for static slides is set to $8$ frames.

\subsection{Qualitative and Quantitative Evaluation}
We evaluate our method using precision, recall, and $F_1$ score of slide transitions for our test dataset. Since the gradual transitions are annotated and predicted by our method as frame intervals, we compare the closest euclidean distances of the start and end points of the predicted and labeled transitions to a threshold of $20$. This comparison is performed bi-directionally and the mutually valid counts determine the number of true positive transitions. 

\begin{table*}[t]
	\caption{Evaluation of precision, recall, and $F_1$ score of slide transition detection for the test set with $14$ videos. In the top rows is the comparison of the first step of the approach: 2-D ResNet18 versus difference with Gaussian blur in both color and grayscale. In the bottom rows the combination of the above methods with the 3-D ResNet50 (in color) and the application of the three networks in reverse order (first 3-D then 2-D) is shown.}\label{comparisonTable}
	\centering
	\begin{tabular}{l@{\hskip 2.0em}*{7}{c}}
	\toprule
 & Number  & TP & FP & FN & \textbf{Precision} & \textbf{Recall} & \textbf{$F_1$ score} \\
 & of transitions &  &  &  &  &  &  \\
	\midrule
Ground Truth & 380 & 380 & 0 & 0 & 100.00 & 100.00 & 100.00 \\ 
\midrule
Diff-RGB-blur & 992 & 365 & 627 & 15 & 36.79 & \textbf{96.05} & 53.21 \\
Diff-gray-blur & 1011 & 365 & 646 & 15 & 36.10 & \textbf{96.05} & 52.48 \\ 
2-D ResNet18-RGB & 1188 & 358 & 830 & 22 & 30.13 & 94.21 & 45.66 \\  
2-D ResNet18-gray & 911 & 355 & 556 & 25 & \textbf{38.97} & 93.42 & \textbf{55.00} \\ 
\midrule
ResNets-Reverse-RGB-gray & 366 & 303 & 63 & 77 & 82.79 & 79.74 & 81.23 \\ 
Diff-RGB-blur + 3-D ResNet50-RGB & 435 & 364 & 71 & 16 & 83.68 & \textbf{95.79} & 89.33 \\ 
Diff-gray-blur + 3-D ResNet50-RGB & 442 & 364 & 78 & 16 & 82.35 & \textbf{95.79} & 88.56 \\ 
SliTraNet-RGB-RGB & 453 & 357 & 96 & 23 & 78.81 & 93.95 & 85.71 \\ 
SliTraNet-gray-RGB & 408 & 354 & 54 & 26 & \textbf{86.76} & 93.16 & \textbf{89.85} \\ 
	\bottomrule
	\end{tabular}
\end{table*}

\begin{figure*}[tb]
\captionsetup[subfigure]{aboveskip=2pt,belowskip=5pt}
\begin{center}
\begin{subfigure}[b]{.99\linewidth}
\centerline{\includegraphics[width=\textwidth]{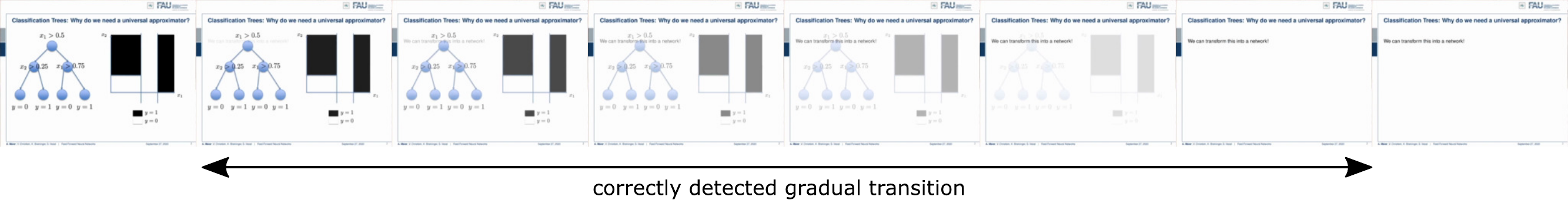}}
\caption{~A true positive gradual transition}\label{fig:qualResults_TP_grad}
\end{subfigure}

\begin{subfigure}[b]{.99\linewidth}
\centerline{\includegraphics[width=\textwidth]{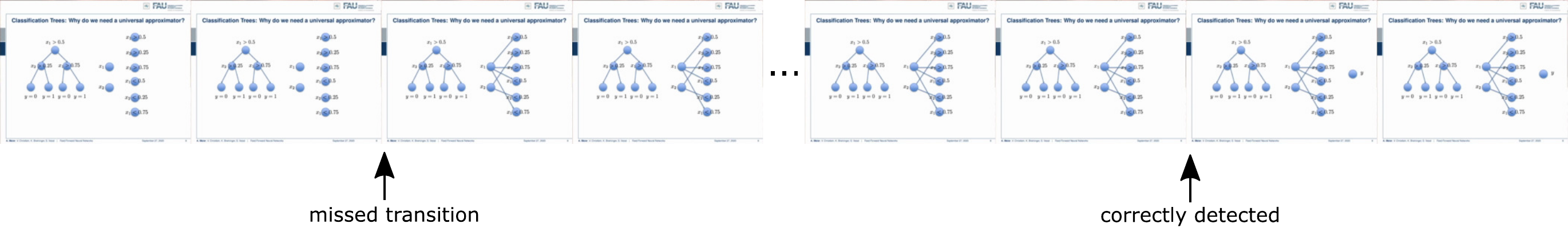}}
\caption{~One false negative transition in an animated slide}\label{fig:qualResults_FP_ani}
\end{subfigure}

\begin{subfigure}[b]{.99\linewidth}
\centerline{\includegraphics[width=\textwidth]{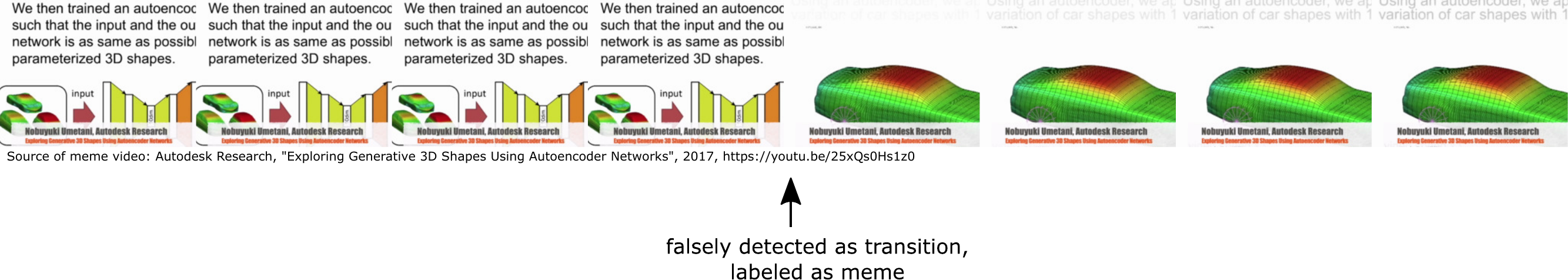}}
\caption{~A false positive transition in case of memes}\label{fig:qualResults_FP_meme}
\end{subfigure}

\begin{subfigure}[b]{.99\linewidth}
\centerline{\includegraphics[width=\textwidth]{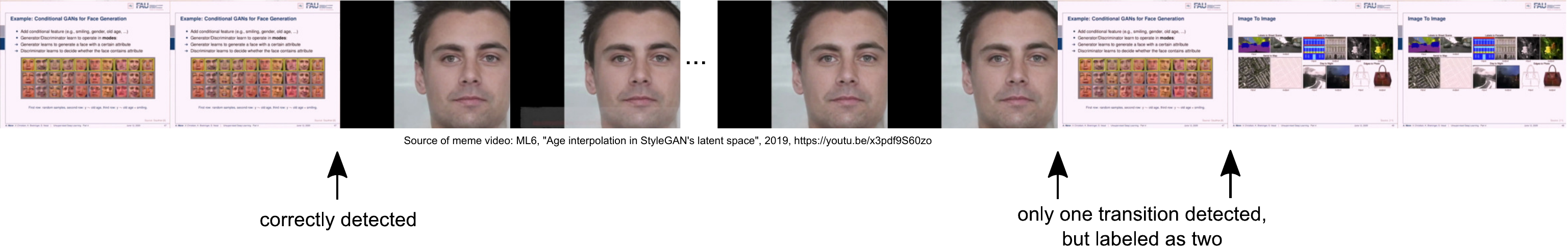}}
\caption{~One false negative transition due to fast slide change after meme insertion}\label{fig:qualResults_FN_meme}
\end{subfigure}
\end{center}
   \caption{Qualitative results for slide transition detection using SliTraNet: Correct and failure cases.} 
\label{fig:qualResults}
\end{figure*}

The quantitative evaluation results are summarized in Tab.~\ref{comparisonTable}. In the top rows, we compare the first step of our approach using the 2-D ResNet18 (trained and tested in RGB and grayscale) to the traditional approach inspired by \cite{PerelmanD2020} of using the frame difference with Gaussian blur (kernel size $k_s=(21,21)$) in RGB and grayscale. From these methods, the grayscale 2-D ResNet achieves the highest $F_1$ score, which is slightly above \SI{50}{\percent}. The 2-D methods have a high recall but a low precision due to their high number of false positives. The frame to anchor comparison detects many false positive transitions for video frames, where short static sequences alternate with motions. 

Hence, the second part of our pipeline is necessary to reduce these false positives, whose results are shown in the bottom rows of Tab.~\ref{comparisonTable}. Using the combination of the first step and the 3-D ResNets a performance gain in the $F_1$ score of up to \SI{35}{\percent} is achieved, i.e., our SliTraNet reaches an $F_1$ score of almost \SI{90}{\percent}, which is closely followed by the combination of difference + 3-D ResNets. 
This second step maintains the high recognition rate while decreasing the number of false positives, resulting in higher precision, which is partly due to the spatio-temporal convolutions in the 3-D ResNets that recognize the different transition types better than the 2-D approach.

Additionally, we evaluate how the order of the networks influences the result by reversing the order. First, we apply the 3-D ResNet to classify overlapping video clips of length $8$ into slide-video, slides and videos. We use the slide-video and slide candidates to apply the next 3-D ResNet to classify the remaining clips into the transition types, static slides and videos. We iterate through the potential transition regions and apply the 2-D ResNet pairwise to localize slide changes. This approach with an $F_1$ score of around \SI{81}{\percent} misses more slide transitions than the competing methods and due to the high complexity in the first two steps consumes a long execution time. In contrast, SliTraNet takes less than $90$ min to process the $190$ min test data.

Overall our SliTraNet demonstrates high effectiveness in the task of slide transition detection in lecture videos, which is also confirmed by the qualitative evaluation. In Fig.~\ref{fig:qualResults} some difficult cases are depicted to highlight the advantage of our method and also define some limitations. One difficulty is represented by animated slides, where little content changes in a short time. Fig.~\ref{fig:qualResults_TP_grad} shows an example of a correctly detected gradual transition, where the start and end point are marked by the blue arrow. From the hard transitions in Fig.~\ref{fig:qualResults_FP_ani} only the right one is detected by SliTraNet. A plausible reason for the failure of the network for the first transition is the small difference of the two frames as 
only thin lines appear that connect the nodes, while for the detected transition the slide change is larger due to the added node. Another difficulty that arises are the memes that are inserted into the lecture videos. The meme in Fig.~\ref{fig:qualResults_FP_meme} has a similar color distribution as the lecture slides and thus the transition within the meme is falsely detected as a slide transition. In Fig.~\ref{fig:qualResults_FN_meme} an example is shown, where the meme was inserted to the end of a static slide.
The slide-video transition is correctly detected, but from the two fast slide changes, only one is detected. In the first step of the approach, we defined that a static slide has to be at least eight frames long, hence slides of one frame length cannot be detected by our method, but for the most applications these limitations are acceptable.

\section{CONCLUSIONS}

We presented a deep learning method to detect slide changes in lecture videos such as hard and gradual transitions. 
The quantitative evaluation showed a high performance of our method for this task.
Future work could comprise extending the approach for a larger dataset and integrating it for online teaching, for instance to automatically insert slides for creating lecture notes in the AutoBlog framework. 

\addtolength{\textheight}{-12cm}   





{\small
\bibliographystyle{IEEEtranS}
\bibliography{ref}
}

\end{document}